\documentclass[conference]{IEEEtran}
\IEEEoverridecommandlockouts
\usepackage{cite}
\usepackage{amsmath,amssymb,amsfonts}
\usepackage{algorithmic}
\usepackage{graphicx}
\usepackage{textcomp}
\usepackage{xcolor}
\usepackage{multirow}
\usepackage{hyperref}
\usepackage{xurl}
\def\BibTeX{{\rm B\kern-.05em{\sc i\kern-.025em b}\kern-.08em
    T\kern-.1667em\lower.7ex\hbox{E}\kern-.125emX}}
\begin{document}

\title{Exploring the Impact of Synthetic Data on Human Gesture Recognition Tasks Using GANs\\}

\author{\IEEEauthorblockN{George Kontogiannis}
\IEEEauthorblockA{\textit{Computer Engineering \& Informatics} \\
\textit{University of Patras}\\
Patras, Greece \\
g.kontogiannis@ac.upatras.gr}
\and
\IEEEauthorblockN{Pantelis Tzamalis}
\IEEEauthorblockA{\textit{Computer Engineering \& Informatics} \\
\textit{University of Patras}\\
Patras, Greece \\
tzamalis@ceid.upatras.gr}
\and
\IEEEauthorblockN{Sotiris Nikoletseas}
\IEEEauthorblockA{\textit{Computer Engineering \& Informatics} \\
\textit{University of Patras}\\
Patras, Greece \\
nikole@cti.gr}
}

\maketitle

\begin{abstract}
In the evolving domain of Human Activity Recognition (HAR) using Internet of Things (IoT) devices, there is an emerging interest in employing Deep Generative Models (DGMs) to address data scarcity, enhance data quality, and improve classification metrics scores. Among these types of models, Generative Adversarial Networks (GANs) have arisen as a powerful tool for generating synthetic data that mimic real-world scenarios with high fidelity. However, Human Gesture Recognition (HGR), a subset of HAR, particularly in healthcare applications, using time series data such as allergic gestures, remains highly unexplored.

In this paper, we examine and evaluate the performance of two GANs in the generation of synthetic gesture motion data that compose a part of an open-source benchmark dataset. The data is related to the disease identification domain and healthcare, specifically to allergic rhinitis. We also focus on these AI models' performance in terms of fidelity, diversity, and privacy. Furthermore, we examine the scenario if the synthetic data can substitute real data, in training scenarios and how well models trained on synthetic data can be generalized for the allergic rhinitis gestures. In our work, these gestures are related to 6-axes accelerometer and gyroscope data, serving as multi-variate time series instances, and retrieved from smart wearable devices. To the best of our knowledge, this study is the first to explore the feasibility of synthesizing motion gestures for allergic rhinitis from wearable IoT device data using Generative Adversarial Networks (GANs) and testing their impact on the generalization of gesture recognition systems. It is worth noting that, even if our method has been applied to a specific category of gestures, it is designed to be generalized and can be deployed also to other motion data in the HGR domain.
\end{abstract}

\begin{IEEEkeywords}
GANs, synthetic data, allergic rhinitis, IoT, gesture recognition, wearable devices, medical data
\end{IEEEkeywords}

% -------------------------------------------------------------
% SECTION
% -------------------------------------------------------------

\section{Introduction}\label{sec:intro}
Over the past few years, research on Human Gesture Recognition (HGR) using wearable devices has gained popularity, mainly attributed to the extensive capabilities unlocked through the application of Machine Learning (ML) and Deep Learning (DL) techniques, some of which are directly deployed to edge devices. HGR technology has been widely applied also in the medical domain (healthcare), providing benefits for addressing health issues that require immediate or costly diagnosis, contributing that way to the field of Healthcare 5.0,  and thereby, enabling real-time medical intelligent passive monitoring. Allergic rhinitis is an example, while attempts have been made for the robust identification of it using wearable devices \cite{aggelides2020gesture}\cite{e2eHGR}, harnessing multi-variate time series IoT data. Usually, when we talk about IoT data, we refer to time series data. A time series, in particular, is simply a quantity of interest observed over a period of time. Multi-variate time series, on the other hand, involve multiple interrelated observations simultaneously over the same time period. However, this kind of data must be well structured and sufficient in quantity to train ML algorithms accurately or to generalize better for disease identification, especially for new users of wearable devices with no historical records \cite{de2023exploring}. For that purpose, a common practice is the usage of data augmentation techniques to generate new data by transforming existing datasets or by generating new synthetic data. In healthcare, data augmentation has been applied, for example, to signals or images to improve disease detection and prediction with the exploitation of GANs \cite{vaccari2021generative}. It is worth mentioning here that the generative models can be manipulated in such a way that they do not contain personal information during the generation, thereby addressing privacy concerns, a significant limitation when utilizing clinical medical data in AI. That said, the applicability of GANs to time series data can solve many issues that current dataset holders face \cite{brophy2023generative}.

Generative Adversarial Networks (GANs) are a class of Machine Learning models consisting of two neural networks, the generator and the discriminator, which are trained together through adversarial training \cite{goodfellow2020generative}. In each learning cycle, the generator generates synthetic data instances from random noise, which is gradually transformed into learned features that mimic the distribution of real data. At the same time, the discriminator learns to distinguish between real and synthetic data, providing feedback to the generator for improvement. Nevertheless, our interest in leveraging GANs for multi-variate time series synthesis encounters several challenges. Firstly, mode collapse is a significant difficulty, where GANs tend to reproduce limited variations of data, failing to capture its diverse modes \cite{goodfellow2020generative}\cite{lin2020using}. Secondly, capturing long-term dependencies within time series data with traditional Multilayer Perceptron (MLP) architectures proves to be difficult, as they struggle to model these extended temporal effects \cite{yoon2019time}\cite{lin2020using}\cite{esteban2017real}. For this reason, in this study, we employ two models that extend the classical architecture of GANs, suitable for time series applications. 

\emph{Our contribution}. In this work, we conduct a study to investigate the feasibility of generative models for the synthesis of multi-variate time series, in the domain of Human Gesture Recognition. Specifically, we leveraged an open-source dataset that contains allergic rhinitis motion gestures retrieved from wearable devices, and we examined if the generated instances meet the three criteria of fidelity, diversity, and generalization. The remainder of the paper is structured as follows: Section \ref{sec:rw} outlines the related work. Section \ref{sec:approach} describes the methodology followed to tackle this AI problem. Section \ref{sec:dataset} provides a quick description of the dataset used. Section \ref{sec:dp} introduces our data processing pipeline steps to prepare our data for ingestion into our AI models. Section \ref{sec:ai} describes our AI modeling, including the mandatory steps followed to train our generative models and the setup for the evaluation process. Section \ref{sec:exps} presents the results of our experiments and their interpretation, while Section \ref{sec:conclusion} concludes the paper.

% -------------------------------------------------------------
% SECTION
% -------------------------------------------------------------

\section{Related Work}\label{sec:rw}
\subsection{Sensor-based Human Gesture Recognition}
Our research group has also contributed to its progress, more specifically with our prior works \cite{aggelides2020gesture}\cite{e2eHGR}. In \cite{aggelides2020gesture}, we extended the methodologies for gesture recognition to address the challenges faced by allergic rhinitis-related gestures, leveraging a wristband Bluetooth device with a 3-axis accelerometer and a 3-axis gyroscope. Particularly, we encountered the gesture recognition problem by exporting a set of features from the triaxial signals of the accelerometer and gyroscope, based on the statistical learning theory and digital signal processing (DSP). These features then fed ML models for training and the classification of the gestures. Building on this foundation, our subsequent work \cite{e2eHGR} introduces an end-to-end framework for the robust and trustworthy identification of allergic rhinitis symptoms, where a modified state-of-the-art Convolutional Neural Network model for time series analysis was established, and data augmentation techniques were utilized only on typical DSP methods (jittering, time warping, magnitude warping, rotation), which, however, pose an extra computational complexity during the training process of the ML models. In addition, our analysis involved the identification and removal of outliers from the dataset to prevent potential biases in the training process of the AI models.

\subsection{Time series GANs}
In the domain of medical time series data synthesis using Generative Adversarial Networks (GANs), RGAN and its conditional counterpart, RCGAN, leverage Recurrent Neural Networks (RNNs) to build both generator and discriminator components, targeting the generation of synthetic medical time series to facilitate the training of supervised models while preserving patient privacy \cite{dash2020medical}. These approaches, however, primarily address the synthesis of uniaxial sensor data, employing temporal features through an RNN-based kernel structure. 

TimeGAN \cite{yoon2019time}, on the other hand, marks a progression by merging the unsupervised capabilities of GANs with the precision of supervised learning, thereby enabling the framework to more accurately represent the dynamics of training data via learned embedding spaces, also utilizing RNN layers. Subsequently, DoppelGANger \cite{lin2020using} emerges as an advancement over TimeGAN, enhancing data fidelity across a wide range of real-world applications. We consider both models in our experiments. 

\subsection{GANs for Sensor Data}
To synthesize sensory data while preserving specific statistical properties of real data, SenseGen was introduced, categorized as a GAN-like architecture \cite{alzantot2017sensegen}. The method involves using both synthetic and real data to train the discriminator, aiming to maintain its accuracy around 50\%. Moreover, SensoryGANs \cite{wang2018sensorygans} first explored the GAN framework's potential for synthesizing motion sensor data to enhance human daily activity recognition models, demonstrating GANs' capability to generate time series data for Wearable-HAR tasks. Additionally, SenseGAN \cite{yao2018sensegan} was developed as a semi-supervised deep learning framework aimed at reducing labeling efforts in Internet of Things (IoT) applications, achieving comparable accuracy to supervised classifiers with only 10\% of the data labeled. Despite their success in generating realistic sensor data for human activities, SensoryGANs require distinct models for different activities, focusing on univariate sensor data without effectively utilizing multi-axial data's temporal and spatial characteristics. Authors in \cite{li2020activitygan}, approached this challenge by utilizing a GAN architecture that merges 1D and 2D CNN layers, focusing on geometric mean reshaping. Regarding the improvement of classification scores and the model's robustness against noisy data, \cite{zhou2021rgc} propose a rotation-based GAN augmentation framework. Authors in \cite{de2023exploring}, examine the impact of synthetic data on HAR classifiers, introducing an evaluation framework for assessing synthetic data quality. For healthcare applications, GANs were employed to generate synthetic vital signs data for COPD patient monitoring, employing validation metrics and Explainable AI (XAI) techniques to ensure data reliability \cite{vaccari2021generative}.
 
\subsection{Evaluation and Quality}
Evaluation and quality assurance in synthetic data generation are pivotal, yet remain among the least resolved aspects of this field due to the absence of universally accepted metrics that can determine the quality of synthetic data. Evaluating the quality of time series data necessitates the need for the temporal correlations between the timesteps to be considered. To overcome these issues, researchers suggested different approaches. In \cite{esteban2017real}, particularly, the authors proposed two evaluation metrics: Train on Synthetic, Test on Real (TSTR), and Train on Real, Test on Synthetic (TRTS). These metrics compute the test accuracy of a machine learning model trained on a set of synthetic data and tested on a set of real data and vice versa. Authors in \cite{figueira2022survey} also suggested simple yet effective methods for evaluating synthetic data by examining fundamental statistics like mean, median, and standard deviation. By comparing these statistical measures with those of the original dataset, the synthetic data can be considered acceptable if the statistics closely match. Additionally, Maximum Mean Discrepancy (MMD) \cite{gretton2006kernel} was employed both in the training and evaluation phases of a GAN \cite{li2015generative} to quantify the similarity between the distributions of real and generated data. Furthermore, the authors of TimeGAN \cite{yoon2019time} utilized the Predictive and Discriminative Score for evaluation, where a 2-layer Long Short-Term Memory (LSTM) network is trained to distinguish between real and synthetic data or to predict the next timestep from the previous points of the time series instances. All the aforementioned metrics except the Predictive Score were used for the evaluation process of our study.

\emph{Our novelty}. Building on the fundamental studies and analyses conducted in our earlier work, our research presents an innovative use of Generative Adversarial Networks (GANs) in the specific context of allergic rhinitis gesture recognition using smartwatch data. Unlike previous studies that mainly focused on general physical activities and specific health conditions like Chronic Obstructive Pulmonary Disease (COPD), our work specifically targets the highly less-explored domain of allergic gestures as well as the domain of Human Gesture Recognition using wearables. We utilize TimeGAN and DoppelGANger models to synthesize high-quality allergic rhinitis gestures and examine if those are suitable for further training on disease identification systems. This approach not only expands the applicability of GANs in medical IoT data but also opens new possibilities in the robust and trustworthy monitoring and analysis of allergic conditions via wearable technology.

% -------------------------------------------------------------
% SECTION
% -------------------------------------------------------------

\section{Approach}\label{sec:approach}
\begin{figure}[htbp]
    \centering
    \includegraphics[width=0.98\columnwidth]{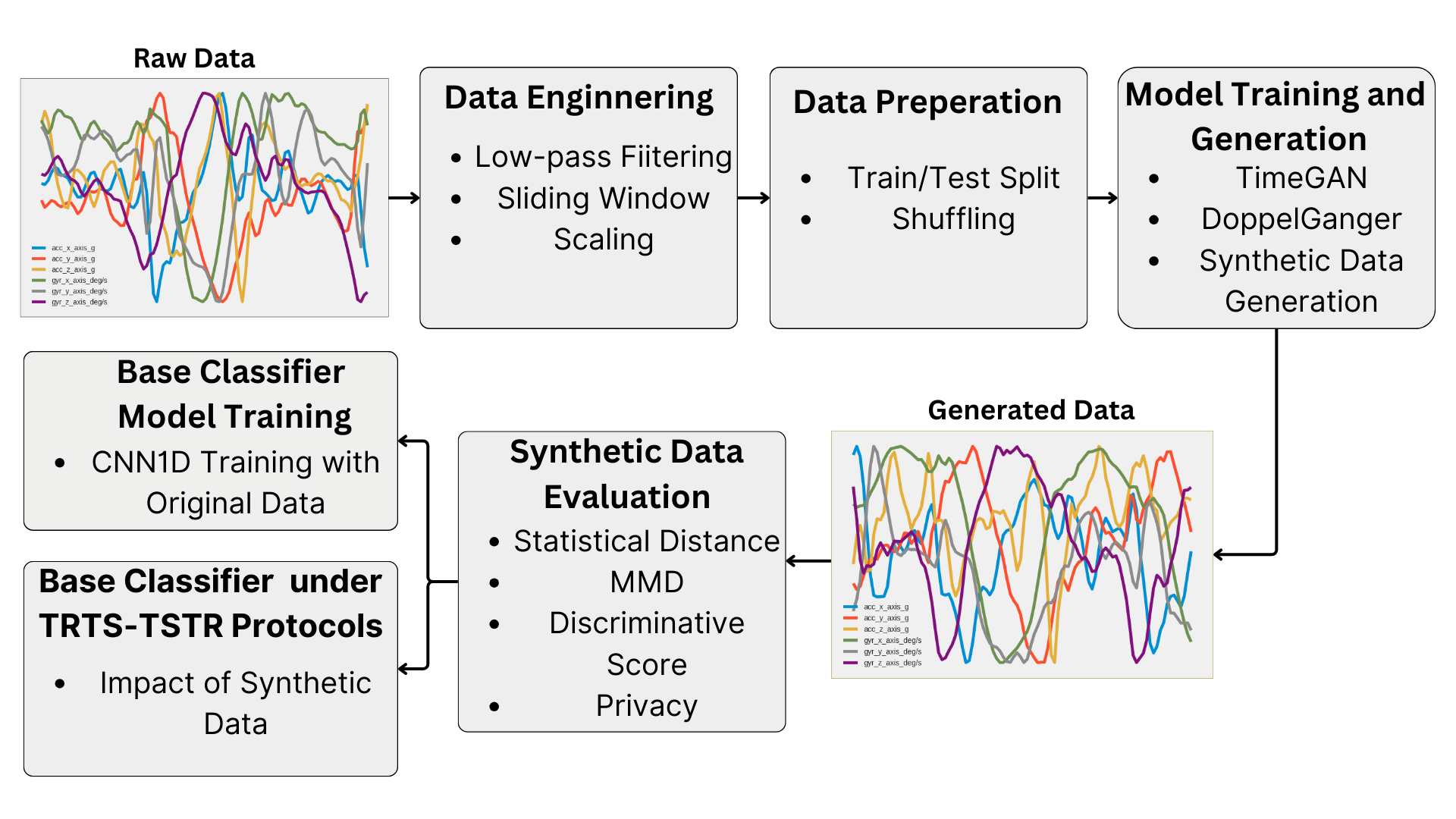}
    \caption{Workflow Diagram.}
    \label{fig:workflow}
\end{figure}

We followed the steps as illustrated in the workflow diagram of Fig.~\ref{fig:workflow}. Firstly, we processed the raw time series instances by applying low-pass filtering and created a held-out set reserved for evaluating the generative models, before any other processing steps. Then we transformed the data into overlapping sequences through the sliding window technique. The instances were then scaled to a specific range of values, shuffled, and split into training and test sets. It is important to note that this test set was utilized only for baseline classification purposes. Regarding the Generative AI, we trained two GAN models, TimeGAN \cite{yoon2019time} and DoppelGanger \cite{lin2020using}, to generate synthetic data, which were evaluated using the five metrics mentioned in Section \ref{sec:rw}, while also by training and testing a 1D convolutional architecture in a TRTS - TSTR setup \cite{dash2020medical}. Finally, for a baseline comparison, we trained the classifier using the real training set and evaluated it on the real test set. This approach allowed us to iteratively refine our models and explore the utility of synthetic data in enhancing machine learning tasks.

\section{Dataset}\label{sec:dataset}
To conduct our study, we exploited a part of the data that comprises an open-source benchmark dataset, which is named GestureSet \cite{10257296}, and has been published by our group\footnote{The dataset can be accessed at: \url{https://www.kaggle.com/datasets/tzamalisp/gestureset-dataset-for-human-gesture-recognition}}.
 The dataset is composed of various gestures that are related to daily activities and the data correspond to accelerometer and gyroscope measurements that originate from wearable devices that embed Inertial Measurement Unit. The dataset conception took place during the life of a national-funded project that was about the intelligent passive identification of allergic rhinitis through the usage of IoT wearable devices and smartphones. For the creation of the specific section of the dataset that we leverage in our current study, 121 individuals (patients) from various allergic clinics participated in the clinical trial for the data collection process, and also an additional 3 from our laboratory, in a more controlled environment \cite{aggelides2020gesture}\cite{e2eHGR}\cite{10257296}. Thus, this section ended up with a collection of 16 distinct classes of allergic gestures, compiled from allergic kinesiological data. 

Regarding the data, each gesture instance is characterized by signals across 6 axes:
\begin{itemize}
    \item Acceleration on the X-Axis, Y-Axis, and Z-Axis, which respectively measure lateral motion or tilt, vertical acceleration or tilt, and forward or backward motion.
    \item Gyroscope Data on the X-Axis, Y-Axis, and Z-Axis, providing insights into the angular velocity around each respective axis.
\end{itemize}
In this study, we exploited the instances of only 4 out of the 16 determined gesture classes, that were collected in a controlled environment in our Laboratory and were identified for their unique patterns that the sensorial data performs:
\begin{itemize}
    \item 01a: The index finger executes a horizontal motion beneath the nose tip
    \item 02a: The hand, shaped into a fist, rubs the eyelids
    \item 03a: The tip of the index or little finger shakes after being inserted into the ear canal
    \item 03b: The earlobe is drawn downwards
\end{itemize}

% -------------------------------------------------------------
% SECTION
% -------------------------------------------------------------

\section{Data Processing}\label{sec:dp}
For our data processing pipeline, we implemented a straightforward yet sufficient approach to prepare the data for model ingestion. At first, we applied a $4th$ order low-pass Butterworth filter with a cut-off frequency of 5Hz to the 6-axes signals of the entire dataset. This step was necessary to reduce both external and device noise and to attenuate abrupt fluctuations. It is worth noting here that, often, removing the gravity component from sensor data is a common practice. However, we chose to retain it in our dataset to evaluate whether our generative models could simulate the effect of gravity on the generated instances. Our goal is mainly to assess the models' capability to reproduce gestures, including the gravity component, thereby testing the AI models' capability to capture the orientation and gravitational influences on motion data. 

In the next step, the dataset underwent sliding window transformation, a basic preprocessing technique to format the input suitably for time series modeling. For this purpose, a window size (\(W_s\)) of 100 timesteps was selected, equivalent to 1 second given a sampling rate of 100Hz, was employed, following the principle of frequency resolution of 1Hz that approximates the movement frequency of human being \cite{hou2019signspeaker}. The overlap (\(op_{step}\)) between consecutive windows was tailored to optimize performance for each model, resulting in distinct configurations: TimeGAN \cite{yoon2019time} seemed to work best with a 99\% overlap, while DoppelGANger \cite{lin2020using} found a 50\% overlap to be more effective. This transformation structured the time series dataset, characterized by the shape \((\text{instances}, \text{timesteps}, \text{features})\), and also augmented the training dataset - a critical aspect of these data-hungry models. Augmentation with overlapping windows enriches the dataset, providing a broader range of patterns for the models to learn, thereby enhancing their ability to capture complex temporal dynamics. 

Moving to the next step, each feature across all timesteps and instances was scaled based on its global minimum and maximum values (min-max scaling), ensuring uniform feature representation throughout the dataset. This normalization is important in the context of AI models when there are significant differences in value ranges across the dataset's features, in our case, accelerometer and gyroscope axes. Without scaling, feature numerical scale differences can skew the model's learning process, biasing features from one sensor over the other. This approach ensures that all signal axes contribute equally. 

Finally, as our last data processing step, we separated the dataset into train-test subsets and shuffled them, minimizing the chance of sequential windows from the same time series instance being adjacent. The test set is only used for the evaluation process of our classifier. For the generative models' evaluation, we use the held-out set preventing the risk of data leakage bias, for a fair and precise comparison of synthetic instances.

% -------------------------------------------------------------
% SECTION
% -------------------------------------------------------------

\section{AI Modeling}\label{sec:ai}
\subsection{Models Training Setup}
\begin{figure}[htbp]
    \centering
    \includegraphics[width=0.94\columnwidth]{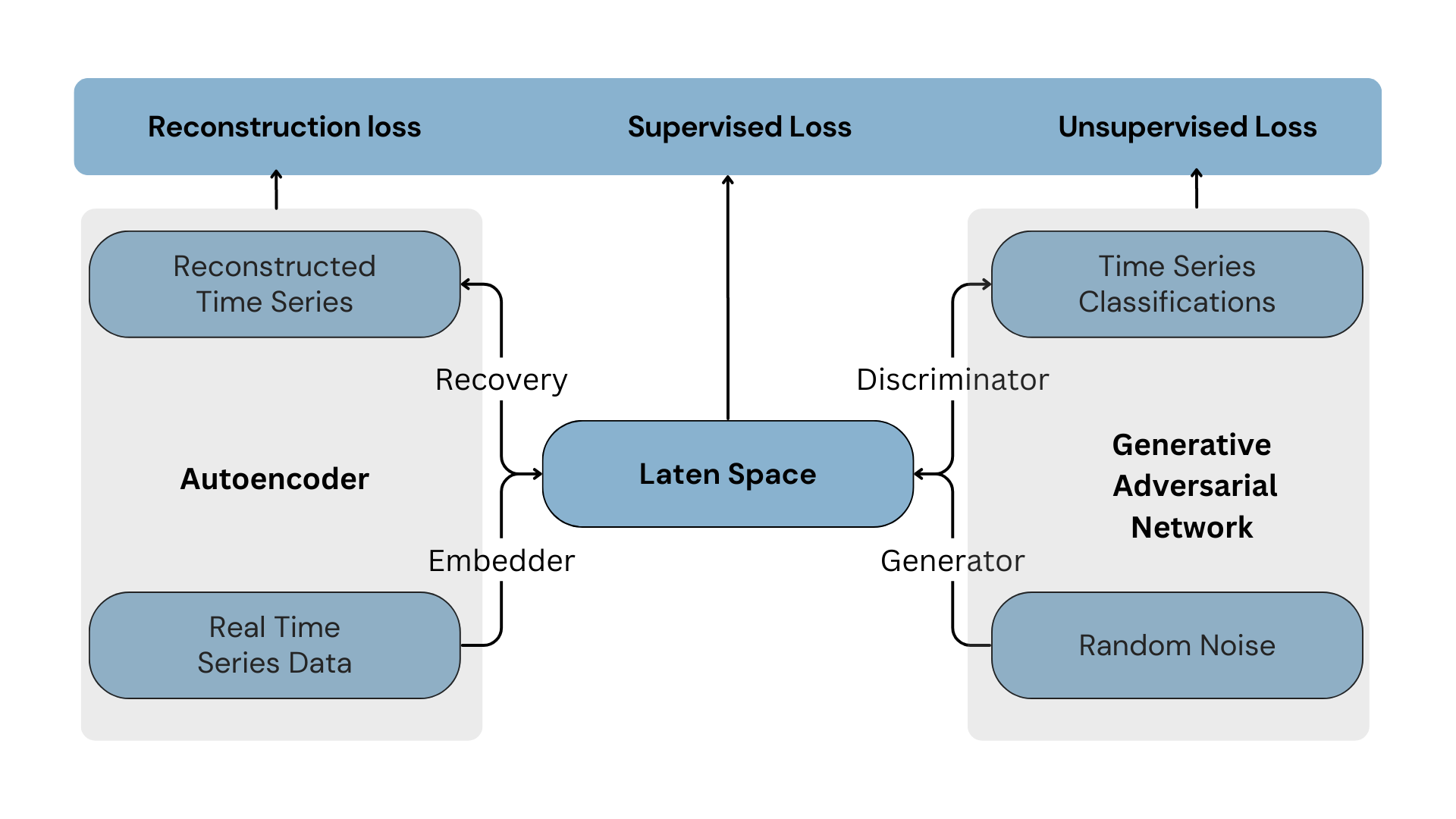}
    \caption{TimeGAN Framework Architecture.}
    \label{fig:timegan}
\end{figure}

\begin{figure}[htbp]
    \centering
    \includegraphics[width=0.99\columnwidth]{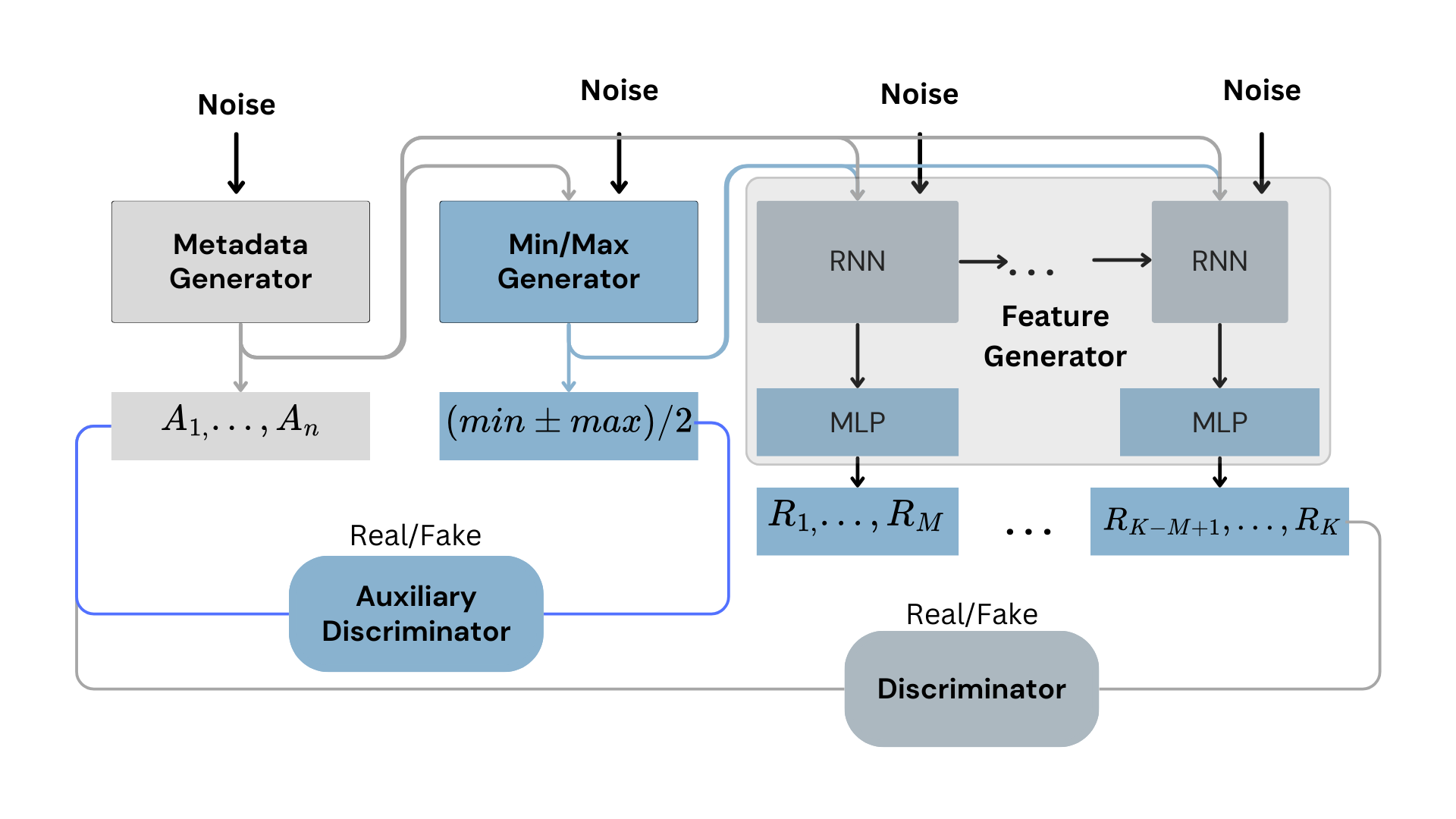}
    \caption{DoppelGANger Framework Architecture.}
    \label{fig:doppelganger}
\end{figure}

According to the AI models that we took advantage of in our study, the TimeGAN \cite{yoon2019time} represents a GAN framework, designed for realistic synthetic sequential data across various domains, by utilizing both supervised and unsupervised learning. In comparison with traditional GAN architectures, such as WGAN \cite{gulrajani2017improved}, TimeGAN introduces an embedding network to reduce the dimensionality of the adversarial learning space and employs a supervised loss to capture the conditional distributions over time. As illustrated in Fig.~\ref{fig:timegan}, the model's architecture is based on three types of losses: the reconstruction loss, the supervised loss, and the unsupervised loss. The reconstruction loss is associated with the autoencoder (embedder and recovery) and evaluates the reconstructed data. The supervised loss refers to the generator's ability to predict the next timestep in the latent space, while the unsupervised loss reflects the adversarial relationship (min-max game) between the generator and the discriminator. TimeGAN's training is separated into three phases: (1) autoencoder training for optimal data reconstruction, (2) supervisor training to learn the temporal dynamics of the data, and (3) combined training to minimize all three losses. In this study, each instance is characterized by the shape \(\textit{(\text{timesteps}, \text{features})}\), and the model is built with GRU units. For the model parameters, we set the sequence length to 100, noise and latent dimension to 64, batch size to 128, learning rate to $5 \times 10^{-4}$, and epochs to 300.

DoppelGANger (DGAN) on the other hand, illustrated in Fig.~\ref{fig:doppelganger}, is a synthetic data generation framework also based on GANs \cite{lin2020using}. To capture temporal correlations, DGAN employs batch generation, generating multiple records simultaneously, helping to better capture long-duration characteristics, instead of single timesteps generation, where temporal dynamics would probably be forgotten. It tackles mode collapse through auto-normalization, normalizing per instance time series signals, and utilizing the min/max values as metadata for improved diversity in generated data. Furthermore, DGAN models the joint distribution between measurements and attributes, using separate generators for each and an auxiliary discriminator to refine attribute generation. For our implementation, the following configuration worked best. We set batch size at 128, a learning rate to $10^{-3}$, betas set to $(0.3, 0.9)$, latent dimension to 32, gradient penalty to 3, packing degree to 1, with 500 epochs, length of each time sequence to 100, a sample length of 10 (the number of timesteps generated at each RNN), rounds per batch of 2, measurement columns adjusted to our dataset's dimensions and attributes columns set to empty list.

\subsection{Models Evaluation}
At this point, the generation of synthetic data\footnote{The data that have been generated by the generative models and used for the experiments can be found here: \url{https://www.kaggle.com/datasets/gkontogiannis/gan-based-synthetic-wearable-gestures}} for each model is completed, however, it is essential to determine whether the these can be used to train the model without introducing bias or negatively affecting its ability to generalize. For that purpose, various approaches have been applied that are related to the employment of algorithms and performance metrics that were presented in Section \ref{sec:rw}.

\textit{Visualization:} We apply Principal Component Analysis (PCA) \cite{wold1987principal} on both the real and synthetic data by collapsing the temporal dimension. Projecting the data in 2-dimensional space helps us visualize the resemblance of the distributions between the real and generated data.

\textit{Statistical Distance:} We examine the synthetic data employing statistical distances, including minimum, maximum, mean, and standard deviation, computed globally, temporally, and per-feature \cite{figueira2022survey}. To encapsulate these statistical values into a single, we compute the discrepancy between the real and synthetic datas' statistical profiles. The discrepancy is quantified using the Euclidean norm, measuring the difference between the statistical summaries of the real and synthetic datasets.

\textit{Maximum Mean Discrepancy (MMD):} To evaluate the dissimilarity between the distributions of synthetic and real data, we computed the maximum mean discrepancy (MMD) \cite{gretton2006kernel} using the Exponentiated Quadratic kernel. This metric provides a measure of the distance between the two distributions, leveraging the kernel trick to effectively capture complex patterns and relationships in the data without explicitly mapping them to a higher-dimensional space.

\textit{Discriminative Score:} For similarity quantification, we train a classification model to distinguish between synthetic and real data as authors suggest in \cite{yoon2019time}. This was done by training the model on a combined set of synthetic and real instances and assessing its classification error on the held-out evaluation set. The architecture of the discriminator was a 2-layer Long Short-Term Memory (LSTM) network with 64 and 128 units respectively, including dropout layers with a rate of 0.3 between them as the fraction of the input units to drop.

\textit{Privacy:} Privacy of generated data is crucial for many applications, especially in medical data-related problems. In this study, the privacy of generated data is calculated by leveraging the precision of a one-class Support Vector Machine (SVM) classifier as proposed by \cite{nikitin2023tsgm}. Privacy is measured as one minus the precision of the SVM's predictions on a hold-out dataset, with values closer to 1 indicating higher privacy levels.

\textit{Train on Real, Test on Synthetic - TRTS:} TRTS setup \cite{esteban2017real} involves evaluating the performance between synthetic data generated by the two models and the original ones. TRTS, basically, works as a measure of realism, since high accuracy on the test set (the synthetic instances) on a well-trained classifier indicates close resemblance. 

\textit{Train on Synthetic, Test on Real - TSTR:} On the other hand, TSTR \cite{esteban2017real} addresses the concerns of how suitable the synthetic instances are and whether the generative model is suffering from mode collapse. Poor TSTR results indicate that the synthetic data lacks the real data's diversity, suggesting that the generative model might be facing issues with mode collapse. 

\begin{figure}[htbp]
    \centering
    \includegraphics[width=0.8\columnwidth]{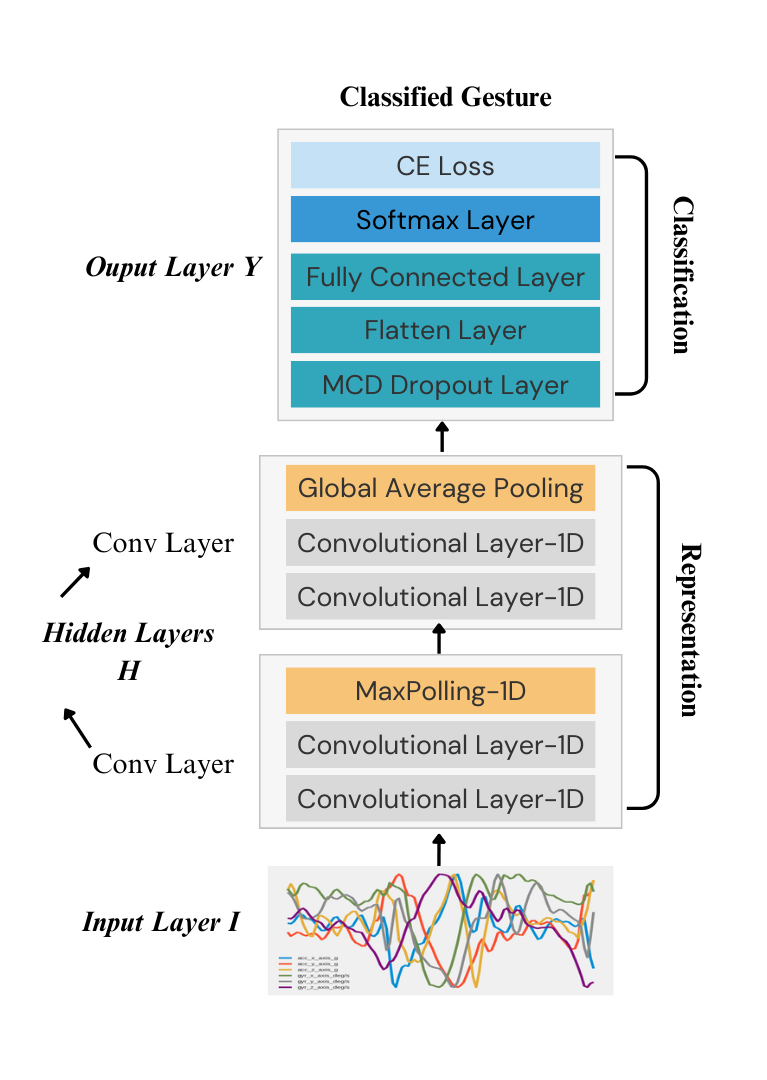}
    \caption{1D Convolution Classifier Architecture.}
    \label{fig:cnn1d}
\end{figure}

For the TRTS - TSTR protocols classifier, we employed a simple 1D architecture composed of Convolutional Layers as shown in Fig.~ \ref{fig:cnn1d}. Specifically, the architecture consists of an input layer, followed by two convolutional layers with a max pooling layer, another two convolutional layers with a global average pooling layer, a Monte Carlo dropout layer with a dropout rate of 0.5, a flattening layer, and a final dense output layer. For the training parameters, the learning rate was set to $1 \times 10^{-4}$, batch size 256, and ran for 30 epochs. The model was evaluated on accuracy, recall, and F1 metrics.

% -------------------------------------------------------------
% SECTION
% -------------------------------------------------------------

\section{Experiments and Discussion}\label{sec:exps}
In healthcare applications, accurately identifying the conditions and the data reliability is more critical than typical classification challenges. For that reason, in our experimental procedure, we planned to accurately determine the quality of the generated instances before using them in the classification step for evaluation. Thus, we evaluated the quality of the generated from AI models' instances based on three essential criteria: a) fidelity, b) diversity, and, c) generalization \cite{alaa2022faithful}\cite{de2023exploring}. That said, we first evaluate the synthetic instances per class since a generative model can capture the underlying characteristics of each class with varying degrees of success. This helps us identify which characteristics are well-replicated and which are not. Then, we proceed with a more global picture by employing TRTS-TSTR, concatenating together the four synthetic classes, and forming that way the complete dataset. 

As a first step of our evaluation process, we compare visually the distributions of the synthetic data, for each class, with the real counterpart, by employing Principal Component Analysis (PCA) and using the first two principal components. For that, we selected a subset of the data, by picking 500 randomly permutated indices to ensure a non-biased selection of instances. \emph{Then, the PCA model was fit exclusively with real data}, ensuring that the transformation of synthetic data aligns with the real data distribution, making fair comparisons.
\begin{figure}[htbp]
    \centering
    \includegraphics[width=0.99\columnwidth]{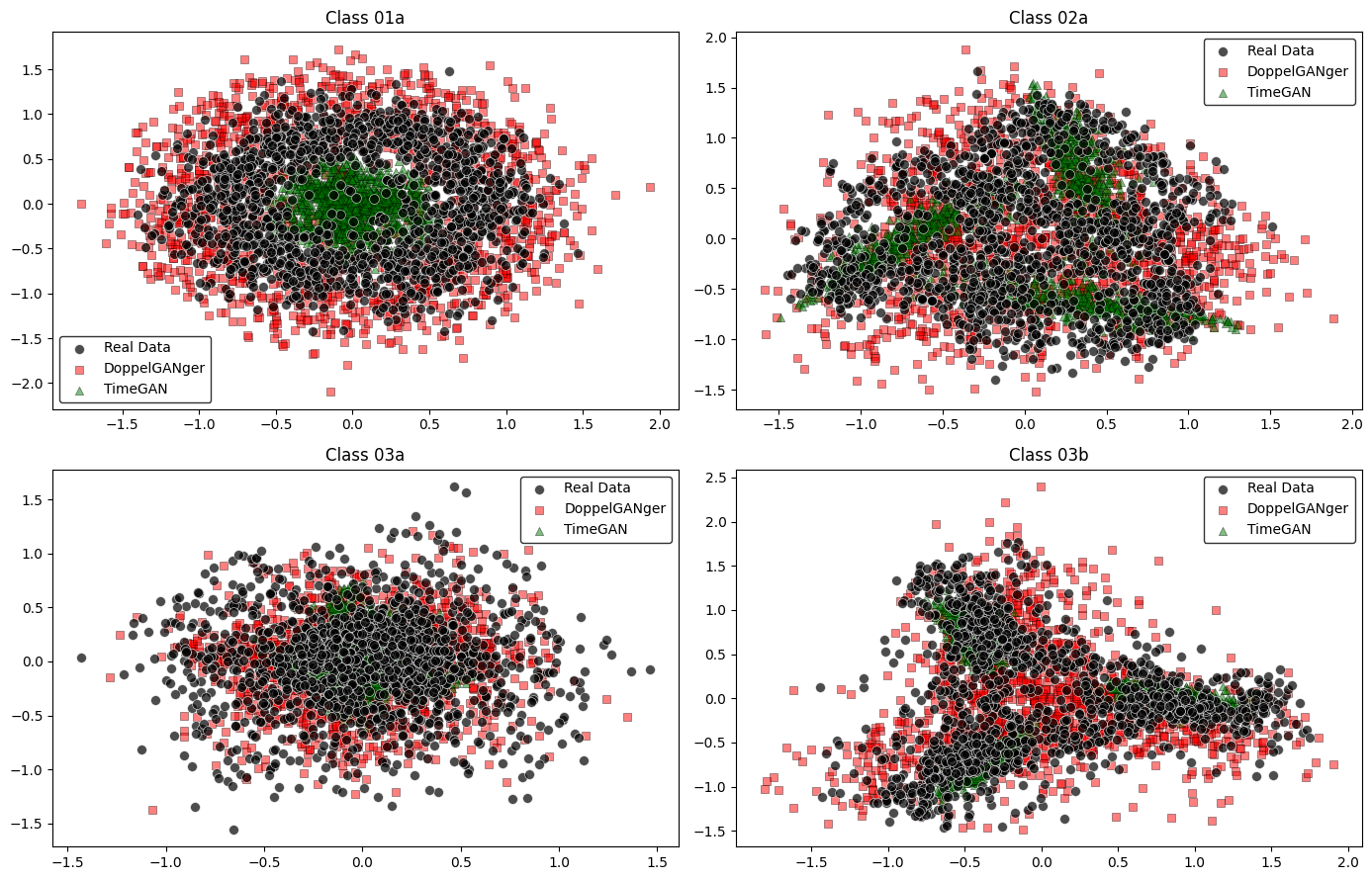}
    \caption{PCA of Real And Synthetic Data Per Gesture.}
    \label{fig:pca}
\end{figure}
In Fig.~\ref{fig:pca}, we present the results of the PCA, which at first sight indicates that both models—DoppelGANger and TimeGAN—are capable of generating all class instances well. In gesture category "01a", the real data points are concentrated around the center formulating a loop shape. DoppelGANger's synthetic data have a broader dispersion that still captures the core pattern of the real data. TimeGAN's synthetic data covers mostly the central region, which may indicate a less precise capture of the underlying data structure. Similar behavior is observed for the category of gestures labeled as "02a". In the case of gesture "03a", the real data presents a more relaxed cluster, and here, DoppelGANger's synthetic data aligns more closely with the real data's dispersion. TimeGAN's output, although overlapping with the real data, shows a more diffuse and less structured distribution, which might indicate limitations in capturing the dynamics of this gesture. The same applies to the category of gestures with the label "03b".

\begin{table}[htbp]
\caption{Evaluation Metrics of Synthetic Data per Gesture}
\centering
\begin{tabular}{|c|c|c|c|c|}
\hline
\textbf{Gest.} & \textbf{Stat. Dist.} & \textbf{MMD} & \textbf{Priv. Score} & \textbf{Disc. Score} \\
\hline
\multicolumn{5}{|c|}{\textbf{DoppelGANger}} \\
\hline
01a & .888 $\pm$ .01 & .005 $\pm$ .001 & .786 $\pm$ .017 & .300 $\pm$ .018 \\
\hline
02a & .803 $\pm$ .02 & .001 $\pm$ .001 & .802 $\pm$ .000 & .154 $\pm$ .024 \\
\hline
03a & 1.164 $\pm$ .03 & .003 $\pm$ .003 & .599 $\pm$ .011 & .219 $\pm$ .016 \\
\hline
03b & .953 $\pm$ .02 & .003 $\pm$ .002 & .690 $\pm$ .030 & .187 $\pm$ .015 \\
\hline
\multicolumn{5}{|c|}{\textbf{TimeGAN}} \\
\hline
01a & 1.710 $\pm$ .01 & .006 $\pm$ .002 & .720 $\pm$ .008 & .38 $\pm$ .016 \\
\hline
02a & 1.246 $\pm$ .015 & .009 $\pm$ .002 & .613 $\pm$ .009 & .28 $\pm$ .017 \\
\hline
03a & 1.913 $\pm$ .03 & .021 $\pm$ .005 & .522 $\pm$ .011 & .42 $\pm$ .025 \\
\hline
03b & 1.168 $\pm$ .022 & .032 $\pm$ .003 & .588 $\pm$ .012 & .42 $\pm$ .014 \\
\hline
\end{tabular}
\label{tab:evaluation_metrics}
\end{table}

Following our evaluation process, we repeat each experiment 10 times to ensure statistical reliability. In Table~\ref{tab:evaluation_metrics}, we present the quantitative metrics of our evaluation experiments, which further amplify the observations from the PCA plots in Fig.~\ref{fig:pca}. For each gesture class, the DoppelGANger model consistently exhibits a lower statistical distance compared to TimeGAN, implying a closer resemblance of the synthetic data to the real. This aligns with the visual interpretation of the scatter points in the PCA distributions. The Maximum Mean Discrepancy (MMD) values, which measure the distance between the distributions of the real and synthetic data, are also smaller for DoppelGANger across all gesture classes. The privacy score, which could reflect the model's ability to generate diverse enough data from the real, ensuring privacy, is generally higher for DoppelGANger. TimeGAN’s lower privacy score might indicate that it replicates the real data at some level, raising potential training and privacy concerns. The discrimination score, which measures how well a discriminator can distinguish between real and synthetic data, is higher for TimeGAN in all gesture classes. This means that it is easier to differentiate between real and synthetic data generated by TimeGAN.

\begin{table}[htbp]
\caption{Results of each model on TRTS and TSTR setup}
\centering
\begin{tabular}{|c|c|c|c|c|}
\hline
\textbf{Eval.} & \textbf{Model} & \textbf{Accuracy} & \textbf{Recall} & \textbf{F1} \\
\hline
 & Base & .880 $\pm$ .031 & .880 $\pm$ .046 & .870 $\pm$ .033 \\
\hline
\multirow{2}{*}{TRTS} & DoppelGANger & \textbf{.848 $\pm$ .027} & \textbf{.845 $\pm$ .021} & \textbf{.848 $\pm$ .029} \\
\cline{2-5}
 & TimeGAN & .823 $\pm$ .048 & .823 $\pm$ .037 & .819 $\pm$ .041 \\
\hline
\multirow{2}{*}{TSTR} & DoppelGANger & \textbf{.873 $\pm$ .012} & \textbf{.875 $\pm$ .010} & \textbf{.873 $\pm$ .012} \\
\cline{2-5}
 & TimeGAN & .815 $\pm$ .047 & .815 $\pm$ .044 & .814 $\pm$ .052 \\
\hline
\end{tabular}
\label{tab:trts_tstr}
\end{table}

For the last step of our evaluation process, we present the performance of each model under the TRTS and TSTR setups in table \ref{tab:trts_tstr}. These experiments help us further understand how well the synthetic data can substitute real data in training scenarios, and how well models trained on synthetically augmented datasets can be generalized. In both the TRTS and TSTR protocols, DoppelGANger outperforms TimeGAN in terms of accuracy, recall, and F1 score, although, all metrics for both models approach the baseline established by training and testing on real data. The higher accuracy and recall indicate that synthetic data enables the model to identify true positives more effectively, and the higher F1 score suggests a balanced precision and recall, which is essential for real-world scenarios. Overall, both DoppelGANger and TimeGAN have proven to be adequate for the synthesis of motion gestures, with DoppelGANger standing out more regarding diversity and generalization while being equally good with TimeGAN at fidelity criterion.

% -------------------------------------------------------------
% SECTION
% -------------------------------------------------------------

\section{Conclusion}\label{sec:conclusion}
In our study, we investigated the use of generative adversarial networks (GANs), and more specifically DoppelGANger and TimeGAN, in the context of allergic rhinitis gesture recognition tasks, using sensory data from wearable devices that compose a part of an open-source benchmark dataset for Human Gesture Recognition (HGR). We focused on the feasibility of the generation of high-quality synthetic data and whether these are suitable for generalization purposes. Our evaluation process provides a picture of each model's capabilities, disadvantages, and room for improvement. More specifically, our findings underscore that both models are adequate for the generation of allergic rhinitis gestures, although with varying degrees of success across critical benchmarks such as fidelity, diversity, and generalization. For future research, we have already started the process of hybrid training, in which synthetic instances are progressively combined with the original dataset, aiming to test the classifier’s ability for better generalization. At the same time, the models' evaluation scores enable us to find the optimal amount of synthetic added instances. Furthermore, instead of using data from only 3 subjects who performed the gestures in a controlled environment, we plan to execute the same AI setup for the data that was collected during the clinical trial, where 121 different subjects participated in random clinical situations, giving that way a variety of how the allergic categories' gestures are performed. Therefore, this approach allows us to further test the capabilities of the models across fidelity, diversity, and generalization. Finally, we are also exploring the integration of Variational Autoencoders (VAEs) and diffusion models into our research pipeline, intending to identify the most effective generative model for allergic rhinitis gesture recognition tasks.

\section{Acknowledgments}
This research has been funded by the Horizon Europe project SynAir-G,
grant agreement ID: 101057271.

% -------------------------------------------------------------
% REFERENCES
% -------------------------------------------------------------
\bibliographystyle{IEEEtran}
\bibliography{references} 

\end{document}